\documentclass[11pt]{article}
\usepackage[final]{acl}

\usepackage{times}
\usepackage{latexsym}
\usepackage[T1]{fontenc}
\usepackage[utf8]{inputenc}
\usepackage{microtype}
\usepackage{inconsolata}
\usepackage{graphicx}
\usepackage{booktabs}
\usepackage{tabularx}
\usepackage{array}
\usepackage{xspace}

\newcommand{\childsafe}{\textsc{ChildSafeAds}\xspace}
\newcommand{\stone}{\textsc{ST1}\xspace}
\newcommand{\sttwo}{\textsc{ST2}\xspace}
\newcommand{\stthree}{\textsc{ST3}\xspace}
\newcommand{\flag}[1]{\texttt{\detokenize{#1}}}

\title{ChildSafeAds Shared Task 2026: Commercial Content in Child-Facing YouTube Videos}

\author{%
  Thales Bertaglia\textsuperscript{1}%
  \thanks{Corresponding author: \texttt{t.f.costabertaglia@uu.nl}} \quad
  Catalina Goanta\textsuperscript{1} \quad
  Gerasimos Spanakis\textsuperscript{2} \quad
  G{\"u}ne{\c{s}} Acar\textsuperscript{3}\\
  \textsuperscript{1}Utrecht University, Netherlands\\
  \textsuperscript{2}Maastricht University, Netherlands\\
  \textsuperscript{3}Radboud University, Netherlands
}

\begin{document}
\maketitle

\begin{abstract}

\childsafe\footnote{The competition is available on CodaBench: \url{https://www.codabench.org/competitions/17595/}.} is a shared task on commercial content in YouTube videos likely to reach children and teenagers. The dataset contains 3,360 videos from 939 channels, each starting with a sponsor segment submitted to SponsorBlock. We pair each segment with its transcript, video, and channel information, plus a sales or service page linked from the video description. Systems determine the type of offer being promoted (\stone), assign product categories (\sttwo), and identify potential legal risks (\stthree). Evidence is divided into four cumulative access levels, from the transcript to the linked page, so performance can be compared against the cost of collecting additional data. YouTube's ``Includes paid promotion'' label was absent from 45.5\% of the videos. GPT-5.4 produced the labels after the organiser team, including a legal expert, reviewed samples and iterated on the taxonomy, prompts and model choices. GPT-5.6-luna independently labelled the development set to measure cross-model agreement. The evaluation received 21 entries; the highest mean macro-F1 was 0.735, while the best transcript-only entry reached 0.652. Across four system papers, the linked promotional page mainly helped identify what was being sold and added little to compliance-risk classification. The competition also exposed sensitivity to rare labels, advertiser overlap across splits and substantial variation in the \stthree labels.

\end{abstract}
\section{Introduction}
Advertising on YouTube often appears inside the video itself, delivered by the same creator and in the same style as the surrounding content. Children and teenagers can struggle to recognise this form of commercial persuasion \citep{de2019influencer,loose2023qualitative}. Disclosure is also inconsistent across countries and platforms \citep{bertaglia2025influencer,gui2024across}. An audit of affiliate-sponsored YouTube videos found a disclosure in about 10\% of cases \citep{mathur2018endorsements}. Authorities seeking to monitor this content at scale cannot begin with disclosures alone.
SponsorBlock provides candidates independent of platform disclosure. It is an open-source crowdsourced browser extension and public API for skipping sponsor segments in YouTube videos.\footnote{\url{https://sponsor.ajay.app/}.} A contributor marks where a sponsorship starts and ends; other users can vote on the submission, and the extension skips accepted segments automatically. The resulting public database provides an independent signal that a video may contain commercial content. YouTube's own ``Includes paid promotion'' label was absent from 45.5\% of the videos in our data, so a monitoring system relying only on that label would miss many relevant candidates.

Once a candidate segment has been found, three questions follow. What is being promoted? What type of product is it? What aspects of the promotion need closer legal review? \childsafe turns these questions into three linked classification tasks. Systems can work from the transcript alone or add video metadata, channel information and the linked promotional page. Every submission records which evidence it used, so the shared task can compare accuracy at different levels of data access and cost.

The evaluation received 21 entries. The highest mean macro-F1 was 0.735, and the best transcript-only system reached 0.652. The four system papers for the task go beyond that leaderboard comparison. In their experiments, the linked promotional page mainly helps identify the offer and product. It adds little to compliance-risk classification, and channel information brings no measurable gain in the systems that test it. Running the competition also exposed weaknesses in the benchmark itself: very rare labels carry a large weight in the metric, advertisers recur across splits, and some compliance-risk labels are unstable.

We first set out the legal context and the tasks, then describe how we built and evaluated the dataset, and close with the participating systems and what they found.

\section{Legal Context}

The labels draw on EU consumer and audiovisual advertising law, limited to what the released data allow us to infer. \stone describes what the consumer receives, using distinctions from the Consumer Rights Directive (CRD) \citep{crd}. \stthree draws mainly on the Unfair Commercial Practices Directive (UCPD), Audiovisual Media Services Directive (AVMSD) and Digital Services Act (DSA) \citep{ucpd,avmsd,dsa}.

The \stthree flags fall into three groups. Disclosure flags ask whether the audience can recognise the content's commercial nature. UCPD Article 5(3) requires the assessment to account for an identifiable vulnerable group, so disclosure clear to an adult may still be inadequate for children. The main legal references are UCPD Articles 6--7 and Annex I point 11, AVMSD Articles 10--11 and 28b, and DSA Articles 26 and 28. We label a promotion as \emph{undisclosed advertising} when the available transcript, description and platform field contain no disclosure. \emph{Inadequate disclosure} applies when a disclosure is present but may not be clear enough for the intended audience.

Content flags cover misleading claims and direct exhortation. UCPD Articles 6--7 address misleading actions and omissions. Annex I point 28 prohibits directly urging children to buy a product or persuade an adult to buy it for them. We use a narrow test: language that pressures a child to make the purchase happen qualifies, while a generic instruction to click a link, download an app or use a code does not. Product flags cover age-restricted or prohibited products and the marketing of foods high in fat, salt or sugar, grounded mainly in AVMSD Article 9 and DSA Article 28.

The released data include files that map each flag to the relevant provisions and explain the connection. They provide citations and notes rather than reproducing the law verbatim. The flags indicate cases for further review. They do not capture every fact needed for a legal assessment, and a flag is not a finding of infringement.

\section{Related Work}

Research on influencer advertising has measured disclosure practices across platforms and countries and studied how children and teenagers recognise commercial persuasion \citep{mathur2018endorsements,bertaglia2025influencer,gui2024across,de2019influencer,loose2023qualitative}. Related work examines children's economic exploitation, adolescent advertising literacy and food marketing \citep{van2020child,zarouali2020adolescents,martinez2021health}. These studies establish why disclosure and product type matter for child-facing content, but they do not provide a benchmark for classifying both from the same videos.

Computational work has detected sponsored YouTube segments and used model explanations to support human labelling and assess influencer-marketing compliance \citep{rodrigues2021identifying,bertaglia2023closing,gui2025evaluating}. \childsafe uses SponsorBlock's crowd-submitted segment boundaries for discovery and focuses on classifying the resulting cases. In legal NLP, LexGLUE benchmarks legal text analysis and legal knowledge \citep{chalkidis2022lexglue}, while consumer-law research connects legal rules to data-driven analysis \citep{lippi2020force}. Recent work also combines legal analysis with platform data to audit DSA transparency \citep{kaushal2024automated} and youth safety \citep{xue2025towards}.

The closest shared-task precedent is LegalLens, which links the detection of legal violations in ordinary text to supporting legal provisions \citep{hagag2024legallens}. \childsafe follows the same structure, connecting evidence from a video, the platform and the promoted product to a set of legally grounded labels.

\section{Task Definition}

Each \childsafe instance starts with a crowd-submitted SponsorBlock interval: a start and end time within a YouTube video where a contributor identified a sponsorship. The dataset also includes the words spoken during that interval, information about the video and channel, and one likely sales or service page linked from the video description. Systems do not decide whether the segment is commercial or whether the channel is child-facing; both are defined when the dataset is built.

The three tasks share the same instances and evidence. Teams may enter any subset. Because every instance begins with a SponsorBlock mark, the benchmark does not evaluate commercial-content detection and contains no non-commercial examples.

\paragraph{ST1: Commercial type.}

\stone asks what the consumer receives through the promoted transaction. It assigns one of five labels: \emph{physical goods}, \emph{digital content or services}, \emph{physical services}, \emph{no identifiable offer}, or \emph{other}.

\paragraph{ST2: Product category.}

\sttwo asks what kind of product is being promoted. It assigns one or more of twelve categories covering apps, hardware and electronics, food, fashion, health, education, financial products, gambling, gambling-adjacent mechanics, toys, creator communities and other products.

\paragraph{ST3: Compliance-risk indicators.}

\stthree asks which aspects of the promotion potentially require closer legal review. It assigns any of six flags: misleading claim, inadequate disclosure, undisclosed advertising, direct exhortation, age-restricted or prohibited product, and marketing for foods high in fat, salt or sugar (HFSS). The labels \emph{no flag} and \emph{insufficient context} stand alone. Systems predict the flags, not the legal provisions behind them. The labelled data include short supporting quotes when the quoted text matches the released evidence.

Figure~\ref{fig:instance} shows how the fields work together. The linked page identifies an online betting service, which determines the product category. The transcript says that a bet cannot lose, supporting the misleading-claim flag. The video did not include a paid-promotion label, but the speaker names the sponsor in the segment, so the instance receives no disclosure flag.

\begin{figure*}[t]

\centering

\setlength{\tabcolsep}{4pt}
\begin{tabularx}{\textwidth}{@{}>{\raggedright\arraybackslash}p{0.20\textwidth}X@{}}
\toprule
\textbf{Record stage} & \textbf{Anonymised content}\\
\midrule
1. Commercial segment & The speaker names [brand] as the sponsor, promotes a sports bet and says that the bet cannot lose.\\[2pt]
2. Description link & \texttt{[URL]}\\[2pt]
3. Resolved page & [page title]; online sports betting and gambling service.\\[2pt]
Platform disclosure & \texttt{false}\\
\midrule
\textbf{Task output} & \textbf{Released label}\\
\midrule
\stone & \flag{digital_content_or_services}\\
\sttwo & \flag{gambling}\\
\stthree & \flag{age_restricted_or_prohibited_product}; \flag{misleading_claim}\\
4. Released evidence & \flag{age_restricted_or_prohibited_product}: ``This episode of [programme] is sponsored by [brand].'' The resolved page identifies an online betting service. \flag{misleading_claim}: ``[brand] is giving you a bet you can't lose.''\\
\bottomrule
\end{tabularx}

\caption{A real training instance with identifiers removed. Source summaries are paraphrased; evidence quotes reproduce the release after redaction and punctuation adjustments. Labels are unchanged.}

\label{fig:instance}

\end{figure*}

\section{Dataset Construction}

\paragraph{Source data and sampling.}

On 5 May 2026, we downloaded a copy of SponsorBlock's public database. We selected intervals submitted in the sponsorship category when they had at least one more upvote than downvote. When a video contained several eligible intervals, we kept the highest-voted one. Contributors make these submissions to improve their viewing experience, not to make a legal determination. We use them only to build the initial pool of likely commercial segments.

We then selected channels likely to reach a substantial teenage audience. The screening combined keywords and other public information about each channel with a GPT-5.4 classifier. It was designed to remove clearly adult-oriented channels, not to estimate each viewer's age. In the released task, ``child-facing'' means that a channel passed this screening process. Systems receive that designation and do not predict it.

\paragraph{Evidence collection.}

For each selected video, we collected the transcript around the marked interval, the video title and description, and basic channel information. We also followed links in the description, resolved redirects and extracted text from the page most likely to describe the promotion. We added a video to the dataset only when at least one link produced a usable page. This requirement removed 1,116 of 4,985 candidates (22.4\%), so the dataset underrepresents expired links and pages that block automated access.

Video descriptions often contain several links. Of the 3,300 instances with a transcript, GPT-5.4 compared the marked segment with the candidate links and selected a likely matching page for 2,092. For the other 1,208, we could not confidently match a link, so we kept the first usable page in the description. Some of these fallback pages may belong to a different promotion in the same video. Another 60 instances have no transcript, but still include the available video, channel and page evidence.

We also recorded whether YouTube reported that the video ``Includes paid promotion''. On 17 July 2026, this field was present for 1,791 videos, absent for 1,529 and unavailable for 40. The 45.5\% absence rate is tied to that collection date because YouTube can change both the label and the interface that exposes it.

\paragraph{Splits.}

The 3,360 instances come from 3,360 videos and 939 channels, split into train (2,353 instances, 632 channels), dev (504, 154) and test (503, 153). No channel appears in more than one split. A system therefore cannot learn a channel's presentation style or disclosure habits from training data and encounter the same channel at test time. Brands and destination pages can recur, so the benchmark tests generalisation to new channels rather than new advertisers.

\paragraph{Data access levels.}

\begin{table*}[t]

\centering

\small

\renewcommand{\arraystretch}{1.12}

\setlength{\tabcolsep}{6pt}

\begin{tabularx}{\textwidth}{clXl}

\toprule

Level & Field group & Contents & Collection cost\\

\midrule

1 & \texttt{transcript} & Segment text, start and end time & Lowest; platform-scale text\\

2 & \texttt{video\_context} & Video ID, title, description, platform disclosure & One metadata call per video\\

3 & \texttt{channel\_context} & Channel ID and channel name & One channel lookup\\

4 & \texttt{product\_page} & Raw and resolved URL, page title and page text & Outbound crawl; pages may be dead or block automation\\

\bottomrule

\end{tabularx}

\caption{Access levels are ordered by the cost of obtaining fields at platform scale. Teams report the highest level used, so accuracy can be interpreted alongside collection cost.}

\label{tab:access}

\end{table*}

Table~\ref{tab:access} groups the released fields by the work needed to collect them. Level 1 requires only the marked transcript. Level 2 adds video metadata, level 3 adds channel information, and level 4 adds the crawled promotional page. The levels are cumulative. Teams report the highest level they use, allowing comparison between systems with different data requirements.

\begin{table*}[t]

\centering

\small

\renewcommand{\arraystretch}{1.12}

\setlength{\tabcolsep}{5pt}

\begin{minipage}[t]{0.43\textwidth}

\vspace{0pt}

\centering

\textbf{ST1: commercial type}\\[4pt]

\begin{tabularx}{\linewidth}{@{}Xrrr@{}}

\toprule

Label & Train & Dev & Test\\

\midrule

Physical goods & 1,125 & 218 & 235\\

Digital content or services & 1,069 & 248 & 225\\

Physical services & 122 & 24 & 33\\

No identifiable offer & 35 & 14 & 9\\

Other & 2 & 0 & 1\\

\bottomrule

\end{tabularx}

\end{minipage}\hfill
\begin{minipage}[t]{0.52\textwidth}

\vspace{0pt}

\centering

\textbf{ST2: product category}\\[4pt]

\begin{tabularx}{\linewidth}{@{}Xrrr@{}}

\toprule

Label & Train & Dev & Test\\

\midrule

Apps & 654 & 171 & 124\\

Hardware and electronics & 516 & 120 & 99\\

Other & 412 & 87 & 97\\

Food & 293 & 49 & 52\\

Creator community & 269 & 47 & 70\\

Fashion & 267 & 50 & 78\\

Health & 264 & 55 & 50\\

Education & 140 & 12 & 18\\

Financial products & 135 & 20 & 56\\

Gambling-adjacent & 92 & 17 & 25\\

Toys & 77 & 8 & 29\\

Gambling & 12 & 5 & 3\\

\bottomrule

\end{tabularx}

\end{minipage}

\caption{ST1 and ST2 label counts by split. ST1 assigns one commercial type to each instance; ST2 may assign multiple product categories.}

\label{tab:st12-distributions}

\end{table*}

\begin{table}[t]

\centering

\small

\renewcommand{\arraystretch}{1.12}

\setlength{\tabcolsep}{4pt}

\begin{tabularx}{\columnwidth}{@{}Xlrrr@{}}

\toprule

ST3 flag & Legal family & Train & Dev & Test\\

\midrule

Misleading claim & Content & 1,277 & 260 & 259\\

Inadequate disclosure & Disclosure & 611 & 118 & 139\\

No flag & Housekeeping & 529 & 127 & 112\\

Undisclosed advertising & Disclosure & 352 & 74 & 89\\

Direct exhortation & Content & 304 & 77 & 71\\

Age-restricted or prohibited product & Product & 59 & 16 & 17\\

HFSS food marketing & Product & 40 & 5 & 5\\

Insufficient context & Housekeeping & 15 & 7 & 6\\

\bottomrule

\end{tabularx}

\caption{ST3 compliance-risk counts by split. ST3 is multi-label; family-level evaluation groups the eight labels into four families: disclosure, content, product and housekeeping.}

\label{tab:st3-distribution}

\end{table}

Tables~\ref{tab:st12-distributions} and~\ref{tab:st3-distribution} show strong label imbalance. The most common \stthree flag is misleading claim, while each product-risk flag has fewer than 60 training examples. The rarest labels are \stone \emph{other} (2), \sttwo \emph{gambling} (12) and \stthree \emph{insufficient context} (15). Macro-F1 weights each label equally, but scores for these rare labels remain sensitive to a small number of predictions.

The release also includes the label taxonomy and legal-reference file. SponsorBlock-derived data retain the CC BY-NC-SA 4.0 licence. The data-use agreement covers video-derived fields, requires compliance with YouTube's Terms of Service, and prohibits redistribution, re-identification, contact and publications that single out a creator.

\section{Annotation and Reliability}

\paragraph{LLM annotation.}

We used GPT-5.4 to produce the labels. For each instance, the model received the available evidence and label definitions and returned the applicable labels. The process was iterative. The organiser team, including a legal expert, reviewed samples, discussed errors and revised the taxonomy, prompts and model choices. This review led, for example, to a narrower definition of direct exhortation and a new annotation pass for that flag. The sample review was not a systematic double-annotation study, so we do not report human inter-annotator agreement.

For \stone and \sttwo, GPT-5.4 received the extracted brand, the linked page title and text, the transcript, and the video description. The main \stthree pass received the same evidence, except for the page title, along with YouTube's paid-promotion field and summaries of the relevant legal provisions. The revised direct-exhortation pass used only the brand and transcript.

To measure label stability across models, GPT-5.6-luna independently labelled the 504 development instances.

\paragraph{Cross-model agreement.}

\label{sec:agreement}

The two models assign the same \stone label in 94.4\% of cases. For multi-label \sttwo, they assign the same label set in 66.3\% of cases; their average set overlap, measured by Jaccard similarity, is 0.74. They assign the same full \stthree flag set in 44.4\% of cases. The largest differences concern inadequate disclosure and misleading claims. The GPT-5.4 labels include inadequate disclosure in 98 cases where GPT-5.6-luna does not, while GPT-5.6-luna assigns misleading claim in 128 cases where GPT-5.4 does not. We report both fine-grained and family-level \stthree scores so that results can also be read at the broader disclosure, content and product levels.

\section{Evaluation Setup}

We hosted the competition on CodaBench, an online platform for running reproducible benchmarks \citep{xu2022codabench}. Teams could enter any subset of the three tasks. A submission contains one machine-readable prediction record for each instance; a missing prediction receives zero credit. The development phase ran from 20 July to 10 August 2026. The evaluation phase ran from 11 to 18 August. The final export contains 21 entries, excluding the organiser baseline.

For each task, we compute F1 for each label that occurs in the gold labels or the predictions and average those labels equally (macro-F1). The overall score is the mean of the three task scores. We also report a family-level \stthree score that groups the eight labels into four families: disclosure, content, product and housekeeping. The organiser baseline predicts the majority class for every instance.

Each submission states the highest data-access level used, which CodaBench displays beside the score. Teams can also report their method, estimated cost and whether they retrieved legal material beyond the references supplied with the task. We do not score legal grounding because the submission format does not support reliable evaluation of retrieved provisions. Section~\ref{sec:levels} instead reports how the system papers used the supplied legal material.

\section{Participating Systems}

\label{sec:systems}

The evaluation phase received 21 entries (Table~\ref{tab:leaderboard}). The highest mean macro-F1 was 0.7347. Fourteen entries used all four access levels, six used the transcript alone, and one used the transcript with video context. The best transcript-only entry placed seventh with 0.6521, 0.083 below the top entry. Since the systems also differ in models, training and validation, this gap cannot be attributed to data access alone. Section~\ref{sec:levels} instead compares access levels within the four systems for which we have detailed reports.

\begin{table*}[t]

\centering

\small

\renewcommand{\arraystretch}{1.05}

\setlength{\tabcolsep}{6pt}

\begin{tabular}{@{}rllrrrrrl@{}}

\toprule

\# & Team & Level & Mean & \stone & \sttwo & \stthree & \stthree fam. & System paper\\

\midrule

1 & abc-abc & 1--4 & \textbf{0.7347} & \textbf{0.8496} & 0.7550 & 0.5996 & 0.6834 & \\

2 & thylinao$^{a}$ & 1--4 & 0.7260 & 0.8049 & 0.7900 & 0.5830 & 0.6620 & \citet{silchenko-2026-thylinao}\\

3 & broxy & 1--4 & 0.7079 & 0.6464 & \textbf{0.8243} & \textbf{0.6530} & \textbf{0.7031} & \citet{steigerwald-etal-2026-nurnberg}\\

4 & runner$^{b}$ & 1--4 & 0.7022 & 0.7923 & 0.7945 & 0.5197 & 0.6359 & \citet{narayan-2026-alphastream}\\

5 & rudolpheric & 1--4 & 0.6699 & 0.6581 & 0.7364 & 0.6153 & 0.6984 & \\

6 & 9999\_9999\_99 & 1--4 & 0.6666 & 0.6093 & 0.8034 & 0.5870 & 0.6688 & \\

7 & Happy & 1 & 0.6521 & 0.7305 & 0.6967 & 0.5290 & 0.6201 & \\

8 & moumita & 1--4 & 0.6261 & 0.6189 & 0.6724 & 0.5870 & 0.6694 & \\

9 & shayyy & 1 & 0.6256 & 0.5778 & 0.7954 & 0.5036 & 0.6176 & \\

10 & k\_n\_n\_nlp & 1 & 0.6252 & 0.5980 & 0.7607 & 0.5170 & 0.6642 & \\

11 & AI-Monitors & 1--4 & 0.6223 & 0.5906 & 0.6979 & 0.5784 & 0.6501 & \citet{sahoo-etal-2026-ai}\\

12 & koshi & 1--4 & 0.6205 & 0.5734 & 0.7233 & 0.5647 & 0.6650 & \\

13 & g\_n\_n\_grease & 1 & 0.5933 & 0.5977 & 0.7221 & 0.4600 & 0.5547 & \\

14 & Abid & 1--4 & 0.5790 & 0.5689 & 0.7559 & 0.4121 & 0.6006 & \\

15 & thenandinee & 1 & 0.5721 & 0.5491 & 0.6818 & 0.4854 & 0.5900 & \\

16 & vimalkumark & 1--4 & 0.5709 & 0.5168 & 0.6885 & 0.5074 & 0.6155 & \\

17 & midushi & 1--4 & 0.5625 & 0.5447 & 0.6696 & 0.4733 & 0.6231 & \\

18 & mercancanpolat & 1--2 & 0.5587 & 0.5203 & 0.6229 & 0.5329 & 0.6692 & \\

19 & taitran501 & 1--4 & 0.5499 & 0.5396 & 0.7090 & 0.4011 & 0.5139 & \\

20 & Axiom & 1--4 & 0.5410 & 0.5964 & 0.6669 & 0.3598 & 0.4880 & \\

21 & avkiran & 1 & 0.4926 & 0.4781 & 0.5571 & 0.4427 & 0.5713 & \\

\midrule

-- & Majority class & -- & 0.0818 & 0.1274 & 0.0330 & 0.0850 & 0.1856 & \\

\bottomrule

\end{tabular}

\caption{Final evaluation-phase leaderboard (503 test instances, macro-F1). The top four entries lie within 0.033 of each other on the mean, and no entry leads all three subtasks. Level is the highest data-access level the team declared. Bold marks the best score in each column. $^{a}$The entry includes three labels set by hand; the system alone scores 0.6537. $^{b}$The entry was chosen between two variants with one leaderboard probe; the probe-free primary system scores 0.6798. Both teams disclosed these choices.}

\label{tab:leaderboard}

\end{table*}

We invited five teams to write system description papers, and four did. Rank was one criterion, but not the main one. The access levels were designed to test a practical question: how much additional data does a monitoring system need, and when is the cost of collecting it worthwhile? Before the deadline, we explicitly told participants that a transcript-only system performing close to one using the full crawl would itself be an interesting result. We therefore also considered the information reported in the submission questionnaire, whether the reported claims were supported by the submitted results, and whether the invited systems represented different approaches. Each invitation asked teams to report data-access levels, validation protocol, test-set cost and negative results.

\paragraph{thylinao.} \citet{silchenko-2026-thylinao} combines six members at equal weight: two fine-tuned encoders, a zero-shot Qwen3-32B, a LoRA-tuned Qwen3-14B used only for \sttwo, a TF-IDF model and a stacker for \stthree. Every threshold is fitted on channel-grouped out-of-fold predictions and never on the development set. The zero-shot member contributes the most while being one of the cheapest components, whereas removing the fine-tuned DeBERTa produces no measurable change. One 14-minute GPU pass dominates inference over the test set.

\paragraph{Nürnberg NLP.} \citet{steigerwald-etal-2026-nurnberg} (entry \texttt{broxy}) build a nine-voter ensemble for each subtask from branches that differ in backbone, adaptation method and class scope. Voters are selected by channel-disjoint cross-validation, with the development set used only as a transfer check. The entry has the highest \sttwo, \stthree and family-level \stthree scores, while its difference from the top overall entry comes entirely from \stone. At full access, it labels the test set in under three GPU-hours on one A100.

\paragraph{Alphastream.} \citet{narayan-2026-alphastream} (entry \texttt{runner}) combines a multi-task ModernBERT-large encoder with an entailment-style reformulation of \stthree and merges their probabilities label by label under a paired bootstrap. The system uses no generative model, and all experiments cost about US\$17 in total. In a single-seed ablation, adding the four access levels increases mean macro-F1 by 0.042, compared with 0.128 from increasing encoder scale. Among five negative results, retraining on 21\% more data lowered the test score, and seed variance was as large as the largest modelling gains.

\paragraph{AI-Monitors.} \citet{sahoo-etal-2026-ai} use different input levels and models for each subtask. A ModernBERT-large encoder reading all four levels handles \stone and \sttwo. For \stthree, a transcript-only encoder is averaged with a QLoRA-tuned Qwen2.5-7B-Instruct for the two disclosure flags. The system labels the test set in 25.5 GPU-minutes. Two of their other submissions ranked above this system on the development set, but the ordering reversed on the test set.

\paragraph{Contribution of the access levels.}

\label{sec:levels}

The four system papers are the best evidence we have on what each additional data source contributes. Their comparisons are not directly equivalent: thylinao ablates its zero-shot member, Nürnberg NLP pools about 150 voters per level, Alphastream runs a single-seed ablation and AI-Monitors compares encoder runs. Even so, all four draw the same line between product identification and compliance-risk classification.

The linked promotional page is most useful for identifying what is being sold. Adding it raises \sttwo by 0.128 for AI-Monitors. For Nürnberg NLP, the level-4 gain is concentrated in \stone (+0.080) and \sttwo (+0.116). The same page contributes little to \stthree: the score falls by 0.031 for AI-Monitors, rises by 0.013 for Nürnberg NLP and does not change measurably for thylinao.

The cheaper levels add little. Neither thylinao nor Nürnberg NLP finds any benefit from the channel name, and for Nürnberg NLP most of the \stthree improvement is already there once video metadata is added. thylinao and AI-Monitors also tested the legal material supplied with the task: giving a prompted model the relevant provisions did no better than giving it the flag definitions alone.

These are comparisons within individual systems, not controlled estimates of each source's value, and the systems use the evidence differently. The pattern is nonetheless consistent across the four papers. Crawling the linked page helps establish what is being promoted but adds little to the compliance-risk labels. For those labels, part of the ceiling may lie in the labels themselves. The two annotating models agree on the full \stthree label set in only 44.4\% of development instances (Section~\ref{sec:agreement}), and Alphastream finds that near-duplicate training examples often carry different flag sets.

\section{Discussion, Limitations and Ethics}
\label{sec:discussion}

The access levels make the cost of collecting evidence part of the evaluation. A transcript can be processed cheaply across many videos. A promotional page must be found, resolved, and crawled, and it may no longer exist or may block automated access. In the system papers, that extra work helped mainly with identifying the offer or product and added little to compliance-risk classification, while the channel name added nothing measurable. A monitoring pipeline may therefore need different evidence depending on what it is trying to establish, rather than collecting every available source for every task.

Running the competition also exposed evaluation weaknesses we had not seen in the dataset alone. Macro-F1 averages only over labels that occur in the gold labels or the predictions, and because \stone \emph{other} has a single test instance, predicting it or not changes about a fifth of the \stone score \citep{silchenko-2026-thylinao,narayan-2026-alphastream}. Publishing the majority-class baseline's per-subtask scores also made it possible to recover a gold class count in closed form, while the leaderboard revealed that \emph{other} occurred in the test labels. 

The channel-disjoint split handles only one kind of dependence. It keeps the same channel out of both training and test data, but advertisers and destination domains can recur. By thylinao's count, 54.5\% of test advertisers, covering 70.4\% of test instances, also appear in the labelled data, and simply copying an advertiser's majority training label scores 0.4649 on the development set \citep{silchenko-2026-thylinao}. At the domain level, 65.0\% of test instances promote a domain seen in training \citep{sahoo-etal-2026-ai}. This does not make the current split invalid: generalising to new channels was its intended purpose. It does mean the benchmark cannot tell us how well a system handles an advertiser it has never seen. A future edition should either hold out advertisers as well or report results separately for seen and unseen advertisers.

Many differences between systems are also small compared with task noise. Near-duplicate training instances agree on the full \stthree flag set only 43--63\% of the time \citep{narayan-2026-alphastream}, close to the 44.4\% exact cross-model agreement on the development set. Alphastream finds seed variance as large as its largest modelling gains \citep{narayan-2026-alphastream}, while Nürnberg NLP finds that the 504 development instances are not enough to reliably separate its strongest candidates \citep{steigerwald-etal-2026-nurnberg}. Small leaderboard differences should therefore be interpreted cautiously, particularly for \stthree. The top four entries are separated by only 0.033 mean macro-F1, and no system leads all three subtasks.

The dataset does not represent all advertising on YouTube. It contains only videos with a SponsorBlock sponsorship mark and a promotional page that we could crawl. Both requirements introduce selection bias, and we have no non-commercial examples to measure discovery at a realistic base rate. The channel screen estimates which channels are likely to reach teenagers from public metadata; it does not measure the actual audience. The selected page could not be aligned confidently to the marked segment in 1,208 instances, and 60 instances have no transcript. Pages can also change after collection.

The benchmark is predominantly English and text-based. It does not include video frames, visual disclosure timing, non-verbal audio or all transaction details that could matter in a legal assessment. These omissions matter most for \stthree. The final labels were produced by GPT-5.4 after the organisers reviewed samples and iterated on the task definitions and prompts. Systems are therefore evaluated against the released taxonomy and model-produced labels, not against an independent legal assessment. The benchmark allows systems to be compared under a common operationalisation of these risks, but its scores should not be read as accurate indicators of legal compliance.

The release contains no viewer records or measured audience demographics, but it includes channel names, video identifiers, titles and descriptions. The data-use agreement prohibits redistribution, re-identification, contact, harassment and publications that single out a creator. We redact identifiers in worked examples. The dataset is intended for research and human review, not profiling or automatic enforcement.

\section{Conclusion}
\childsafe provides a 3,360-video benchmark for classifying commercial offers, product categories and potential compliance risks in child-facing YouTube content. The shared task received 21 entries. In the four system papers, additional evidence helps some tasks and not others: crawling the promotional page improves product identification but adds little to compliance-risk classification, and the channel name brings no measurable gain. For monitoring in practice, this argues for collecting evidence according to the question being asked rather than gathering every source for every task. The competition also showed where the benchmark needs work before a second edition: label stability, advertiser overlap, and handling rare labels in the metric.

\noindent\textbf{Acknowledgements.} We thank the SponsorBlock contributors and the participating teams. This research was supported by the ERC Starting Grant HUMANads (ERC-2021-StG No 101041824).

\bibliography{references}

\end{document}